\documentclass[letterpaper,10pt,conference]{ieeeconf}

\IEEEoverridecommandlockouts
\let\proof\relax
\let\endproof\relax
\relax

\usepackage{algorithm}
\usepackage[noend]{algpseudocode}
\usepackage{amsmath}
\usepackage{amsthm}
\usepackage{amssymb}
\usepackage{amsfonts}
\usepackage{booktabs}
\usepackage{graphicx}
\usepackage{multirow}
\usepackage{tabularx}
\usepackage{times}
\usepackage[hyphens]{url}

\newtheorem{definition}{Definition}
\newtheorem{proposition}{Proposition}

\newenvironment{proofsketch}{%
    \proof}{\endproof}

\title{\LARGE \bf Agent-Aware State Estimation in Autonomous Vehicles}

\author{Shane Parr$^{1*}$, Ishan Khatri$^{1*}$, Justin Svegliato$^{2}$, and Shlomo Zilberstein$^{2}$%
\thanks{This work was supported in part by the NSF grants IIS-1724101, IIS-1813490, and DGE-1451512.}%
\thanks{*Both authors contributed equally.}%
\thanks{$^{1}$Work done while students at University of Massachusetts Amherst. Emails: {\tt\small shane@sparr.io, ishan@khatri.io}}%

\thanks{$^{2}$College  of  Information  and  Computer  Sciences,  University  of  Massachusetts Amherst, MA, USA. Emails: {\tt\small \{jsvegliato, shlomo\}@cs.umass.edu }}%
}

\begin{document}

\maketitle
\thispagestyle{empty}
\pagestyle{empty}

\begin{abstract}
    Autonomous systems often operate in environments where the behavior of multiple agents is coordinated by a shared global state. Reliable estimation of the global state is thus critical for successfully operating in a multi-agent setting. We introduce \emph{agent-aware state estimation}---a framework for calculating indirect estimations of state given observations of the behavior of other agents in the environment. We also introduce \emph{transition-independent agent-aware state estimation}---a tractable class of agent-aware state estimation---and show that it allows the speed of inference to scale linearly with the number of agents in the environment. As an example, we model traffic light classification in instances of complete loss of direct observation. By taking into account observations of vehicular behavior from multiple directions of traffic, our approach exhibits accuracy higher than that of existing traffic light-only HMM methods on a real-world autonomous vehicle data set under a variety of simulated occlusion scenarios.
\end{abstract}

\section{Introduction}\label{sec:intro}

Autonomous systems often operate in environments where the behavior of agents is governed by both their \emph{local state} and a \emph{shared global state} which encapsulates important aspects of the environment. Reliable estimation of the global state is critical to robust multi-agent systems. Centrally managed coordination signals, such as traffic lights for autonomous vehicles~\cite{broggi2012vislab,svegliato2019belief,basich2020learning}, radio beacons for autonomous aerial robots~\cite{achtelik2012sfly,cliff2015online}, and distress signals for search and rescue operations~\cite{goodrich2008supporting,pineda2016continual}, are particularly important examples of shared global state. Improperly receiving or interpreting these signals can cause catastrophic failures.

There are several methods used to mitigate such catastrophic failures. 1) Systems may choose to transfer partial control to a human assistant~\cite{wray2016hierarchical,zilberstein2015building}. 2) Systems may choose to gather additional information to reduce uncertainty~\cite{kaelbling1998planning}. 3) Systems may choose to use limited communication to share critical information between agents to mitigate risk~\cite{goldman2003optimizing}. In high-stakes, partially observable environments where human intervention, information gathering, and collaboration are limited, these methods alone are insufficient. State estimation \emph{must} be robust enough to recover a global state when direct perception fails, thus motivating the goal of our work.

\begin{figure}[t]
    \centering
    \includegraphics[width=\columnwidth]{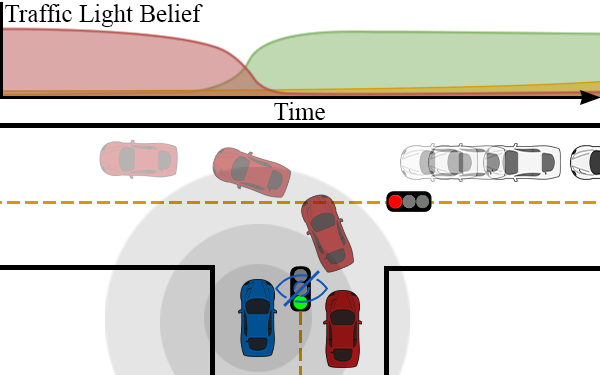}
    \caption{By observing intelligent agents in the environment, an autonomous system can recover coordination signals despite prolonged total obstruction.}
    \label{fig:teaser}
\end{figure}

Figure~\ref{fig:teaser} depicts an example scenario. A blue autonomous vehicle, modeled as an \emph{external observer}, is approaching a three-way intersection with two other agents: the red and white human-operated vehicles. Due to glare on its sensors, the blue vehicle cannot observe the traffic light, a crucial element of the global state. However, by observing the red vehicle turning and the white vehicle stopping, the blue vehicle can properly recover the traffic light's state without direct observation or vehicle-vehicle communication. 

A simple approach to state estimation is to use a Hidden Markov Model (HMM) or other graphical model to smooth missing or erroneous observations. While HMMs can smooth a signal in the presence of noise, they struggle with total obstruction over long periods of time~\cite{vaseghi1997noise}. They also struggle  to incorporate complex information because, in general, the number of states scales exponentially in the number of variables~\cite{ghahramani2001introduction,rabiner1986introduction}, and thus most modeling attempts avoid using additional information to represent the behavior of other agents. However, because HMMs work well for temporal smoothing, they have been widely used in multi-agent scenarios such as traffic light classification~\cite{dymarski2011hidden, gomez2014traffic}. Dynamic Bayesian Networks (DBN) are a common generalization of HMMs and Bayesian Networks that add an internal Bayesian Network structure to the HMM. The structure of DBNs must be given or learned and exact inference may not be feasible in real-time~\cite{murphy2002dynamic}. 

Alternatively, state estimation and decision making can be handled concurrently. A Partially Observable Markov Decision Process (POMDP)~\cite{kaelbling1998planning} operates by maintaining and updating a belief over states and can be extended to multiple decentralized agents as a Decentralized POMDP (Dec-POMDP)~\cite{oliehoek2016concise}. POMDPs and Dec-POMDPs, respectively being NP and NEXP hard to solve optimally~\cite{mundhenk2000complexity, bernstein2002complexity}, are often intractable to use without simplifying assumptions and an accurate joint model of all plausible agent interactions.

In this work we propose a novel framework for \emph{agent-aware state estimation} where noisy observations of state can be combined with direct observations of other agents as they execute a policy. The \emph{agent-aware state estimation} framework allows us to combine the benefits of an HMM with additional information (policies that solve a Dec-MDP) via a DBN representation to reconstruct the global coordination signal in a robust, decision-theoretic manner.

Our key contributions include formalizing a general framework for \emph{agent-aware state estimation} and highlighting the benefits of the transition-independence property. We provide a DBN construction (Figure~\ref{fig:dbn}) for which inference is equivalent to solving the \emph{transition-independent agent aware state estimation problem}. We apply this approach to the problem of traffic light classification by an autonomous vehicle when the traffic light may be obscured. We offer results on a real-world data set of traffic light scenarios, where the behavior of other vehicles from multiple directions of traffic is used to gain information about the state of the intersection. Our results show that our approach both exhibits higher accuracy and is more robust than competing approaches that only use direct image data of the traffic light.

\section{Background}

We base our representation of an agent-aware state estimation problem on a \emph{factored $n$-agent decentralized Markov decision process} (Dec-MDP), a formal decision making framework closely related to Dec-POMDPs. Factored $n$-agent Dec-MDPs model multi-agent cooperative problem-solving in partially observable, stochastic environments using a factored state representation, with the special property that no part of the environment is hidden from every agent~\cite{bernstein2002complexity,amato2013decentralized}. When combined with a known policy, such frameworks allows for accurate and efficient state estimation.

A factored $n$-agent Dec-POMDP is defined by a tuple $\langle \mathcal{I}, \mathcal{S}, \mathcal{A}, \mathcal{T}, \mathcal{R}, \Omega, O \rangle$. $\mathcal{I}$ is a set of $n$ agents. $\mathcal{S} = \mathcal{S}_0 \times \mathcal{S}_1 \times \dots \times \mathcal{S}_n$ is a finite set of factored states: a global state space $\mathcal{S}_0$ and a local state space $\mathcal{S}_{i > 0}$ for each agent $i  \in \mathcal{I}$. $\mathcal{A} = \mathcal{A}_1 \times \dots \times \mathcal{A}_n$ is a finite set of joint actions: an action set $\mathcal{A}_i$ for each agent $i \in \mathcal{I}$. $\mathcal{T}: \mathcal{S} \times \mathcal{A} \times \mathcal{S} \rightarrow [0,1]$ is a transition function that represents the probability $\mathcal{T}(s'|s, \vec{a}) = Pr(s'|s, \vec{a})$ of reaching factored state $s' \in \mathcal{S}$ after performing joint action $\vec{a} \in \mathcal{A}$ in factored state $s \in \mathcal{S}$. $\mathcal{R}: \mathcal{S} \times \mathcal{A} \times \mathcal{S} \rightarrow \mathbb{R}$ is a reward function that represents the expected immediate reward $R(s, \vec{a}, s')$ of reaching factored state $s' \in \mathcal{S}$ after performing joint action $\vec{a} \in \mathcal{A}$ in factored state $s \in \mathcal{S}$. $\Omega = \Omega_1 \times \dots \times \Omega_n$ is a finite set of joint observations: an observation set $\Omega_i$ for each agent $i \in \mathcal{I}$. $O: \mathcal{A} \times \mathcal{S} \times \Omega \rightarrow [0,1]$ is an observation function that represents the probability $O(\vec{\omega}|\vec{a}, s) = Pr(\vec{\omega}|\vec{a}, s)$ of joint observation $\vec{\omega} \in \Omega$ after taking joint action $\vec{a} \in \mathcal{A}$ and ending up in factored state $s \in \mathcal{S}$. A Dec-MDP is a specific class of Dec-POMDP where for each \textit{joint observation} $\vec{\omega} \in \Omega$, there exists some $s \in \mathcal{S}$ such that $Pr(s|\vec{\omega}) = 1$.

A factored $n$-agent Dec-MDP is \emph{locally fully observable} if, for each local observation $\omega_i \in \Omega_i$ and global state $s_0$, there exists some local state $s_i$ such that $Pr(s_i,s_0|\omega_i) = 1$. A factored $n$-agent Dec-MDP is \emph{transition independent} if the transition function $\mathcal{T}$ can be represented by a tuple of transition probabilities $(\mathcal{T}_0, \mathcal{T}_1, \dots, \mathcal{T}_n)$ such that $\mathcal{T}(s, \vec{a}, s') = \mathcal{T}_0(s_0'|s_0) \prod_{i=1}^n \mathcal{T}_i(s_i'|s_i, a_i)$. Transition independent Dec-MDPs are simpler because they obviate the need to use complex joint models. They can approximate situations with weakly-coupled agents~\cite{SZ:BZLGjair04}.

\section{Agent-Aware State Estimation}

Informally, an \emph{agent-aware state estimation problem} (AASEP) is a problem where an \emph{external observer} must estimate state given observations of other agents as they execute a policy to solve a decentralized decision problem. The external observer receives noisy observations of the state space that is factored into two components: 1) the global state, whose transitions are unaffected by the agents and 2) the joint local state, whose transitions are based on both the global state, itself, and the actions of each agent. These problems, called \emph{signalized} problems, (Definition~\ref{def:signalized}) possess one-way independence as the global state acts as a coordinating signal, influencing the actions of the agents, while itself transitioning independently.

\begin{definition}
   Given a factored state space $\mathcal{S} = \mathcal{S}_0 \times \mathcal{S_N}$, where $\mathcal{S}_0$ refers to the \textbf{global state space}, and $\mathcal{S_N} = \mathcal{S}_1 \times \dots \times \mathcal{S}_n$ refers to the \textbf{joint local state space}, a decision problem is \textbf{signalized} if for every state $s \in \mathcal{S}$, joint action $\vec{a} \in \mathcal{A}$, and successor state $s' \in \mathcal{S}$, the transition function $\mathcal{T}$ can be written as a tuple of transition probabilities $(\mathcal{T}_0, \mathcal{T_N})$ such that $\mathcal{T}(s'|s,\vec{a}) = \mathcal{T}_0(s_0'|s_0)\mathcal{T_N}(s_\mathcal{N}'|s_\mathcal{N}, s_0, \vec{a})$.
   \label{def:signalized}
\end{definition}

We now provide a more formal description of the AASEP. 
First, the policies executed by the observed agents are a solution to a signalized Dec-MDP, with the external observer not included in the Dec-MDP. The choice of Dec-MDP is not inherent to our approach; we also explore simpler models.

Second, we define a joint observation function $\mathcal{Z}$ and the corresponding set of possible joint observations $\mathcal{O}$, accessible only to the external observer and separate from definitions internal to the signalized Dec-MDP (e.g.\ $O$ or $\Omega$). Thus, the observer is modeled as receiving a sequence of observations of each state factor as agents execute their individual policies. The goal of the external observer (and thus the objective of the AASEP) is to estimate the current state given the history of noisy observations and their individual policies. A full definition of the AASEP is provided in Definition~\ref{def:AASEP}.

\begin{definition}
    An \textbf{agent-aware state estimation problem} (AASEP) is defined by the tuple $\langle \mathcal{M}, \Pi, \mathcal{O}, \mathcal{Z} \rangle$, where:
    \begin{itemize}
        \item $\mathcal{M}$ is a signalized Dec-MDP: $\langle \mathcal{I, S, A, T, \ \cdot \ }, \Omega, O \rangle$.
        \item $\Pi = \{ \pi_1, \dots, \pi_n \}$ is a set of stochastic policy trees $\pi_i : \Omega^t \times \mathcal{A}_i \rightarrow [0, 1]$ such that $\pi_i(\vec{h}, a_i) = Pr(a_i|\vec{h})$ defines the probability that agent $i$ takes action $a_i$ with history $\vec{h}$ of $t$ observations $\omega_1, \omega_2, \dots, \omega_t$ in $\mathcal{M}$.
        \item $\mathcal{O} = \mathcal{O}_0 \times \mathcal{O}_1 \times \dots \times \mathcal{O}_n$ is a finite set of joint observations: $\mathcal{O}_0$ are observations of a global state $s_0$ and $\mathcal{O}_{i > 0}$ are observations of a local state $s_i$ made by the external observer.
        \item $\mathcal{Z} = \{ \mathcal{Z}_0, \mathcal{Z}_1, \dots, \mathcal{Z}_n \}$ is a set of observation functions $\mathcal{Z}_i: \mathcal{S}_i \times \mathcal{O}_i \rightarrow [0, 1]$ such that $\mathcal{Z}_i(s_i, o_i) = Pr(o_i|s_i)$ defines the probability that the external observer receives $o_i \in \mathcal{O}_{i}$ from state factor $s_i$ for each agent $i$.
    \end{itemize}
    \label{def:AASEP}
\end{definition}

In an AASEP, the observed agents have the potential for complex interactions that provide negligible information about global state. Additionally, to perform exact inference and compute a solution, the external observer must form a belief over the possible beliefs of the agents internal to the Dec-MDP, who are themselves responding to private observation histories $\vec{h}$. For these reasons, approaching the traffic light problem as an unstructured AASEP is impractical, and could come with significant computational complexity. While AASEPs are quite general and worthy of further study, for the remainder of this work we focus on a subclass of problems that assumes full local observability and transition-independence between agents: Transition-Independent AASEPs (TI-AASEPs).

By assuming 1) different local states are transition-independent from each other but not from global state and 2) agents can be well modeled with full knowledge of global and local state, a TI-AASEP can be solved without modeling complex agent interactions or estimating the beliefs of other agents. We work to ensure these assumptions are met for all agents included, but previous Dec-MDP literature suggests that minor violations of transition-independence are handled well in practice~\cite{SZ:BZLGjair04}. These assumptions allow us to specify a TI-AASEP, formally defined in Definition~\ref{def:TIAASEP}.

\begin{definition}
    A \textbf{transition-independent agent-aware state estimation problem} (TI-AASEP) is an AASEP $\langle \mathcal{M}, \Pi, \mathcal{O, Z}\rangle$, where $\mathcal{M}$ is transition-independent and locally fully observable, and $\Pi$ is composed of policies that map from $\mathcal{S}_0 \times \mathcal{S}_i \times \mathcal{A}_i \rightarrow [0,1]$ such that $\pi_i(s_0,s_i,a_i) = Pr(a_i|s_0, s_i)$. 
    \label{def:TIAASEP}
\end{definition}

A TI-AASEP resolves many of the difficulties of an AASEP. By modeling agents as though they have local full observation, we can use a set of Markovian policies (exchanging any $\omega_i$ or $\vec{h}$ with $s_0$ and $s_i$) to approximate agent behavior. Additionally, the transition independence of the Dec-MDP allows for considering the agents as nearly totally separated, joined only by a shared global state.

Solving a TI-AASEP is equivalent to performing inference with a tractable DBN--- one with discrete variables and a constant-sized conditional probability table at each node. External observations $\mathcal{O}_0$ through $\mathcal{O}_n$ form evidence nodes, and the global state space state factor $\mathcal{S}_0$ is our target for inference. The probabilistic relationships between the evidence nodes $\mathcal{O}$ and global state $\mathcal{S}_0$ are characterized through the observation functions $\mathcal{Z}$, transition functions $\mathcal{T}$, and stochastic agent policies $\Pi$.

Figure 2 shows the two-slice DBN representation of a transition-independent AASEP (TI-AASEP). The \emph{purple} nodes represent an ordinary signal estimation HMM, which models a global state $s_0$'s relationship with an external observer's observation $o_0 \in \mathcal{O}_0$. In \emph{blue}, we have the standard task of observing the state of agents in our environment. In \emph{red}, we have the agent model, or action node, which relates the two tasks by depending on both prior local state and global state (via the agent's policy, $\pi_i$). Due to the causal dependence of all $\mathcal{O}_i$ on $\mathcal{S}_0$, we can improve our estimate of $\mathcal{S}_0$ with far more data than if we were restricted to only $\mathcal{O}_0$, direct external observations of global state.

\begin{figure}[t]
    \centering
    \includegraphics[height=2.75in]{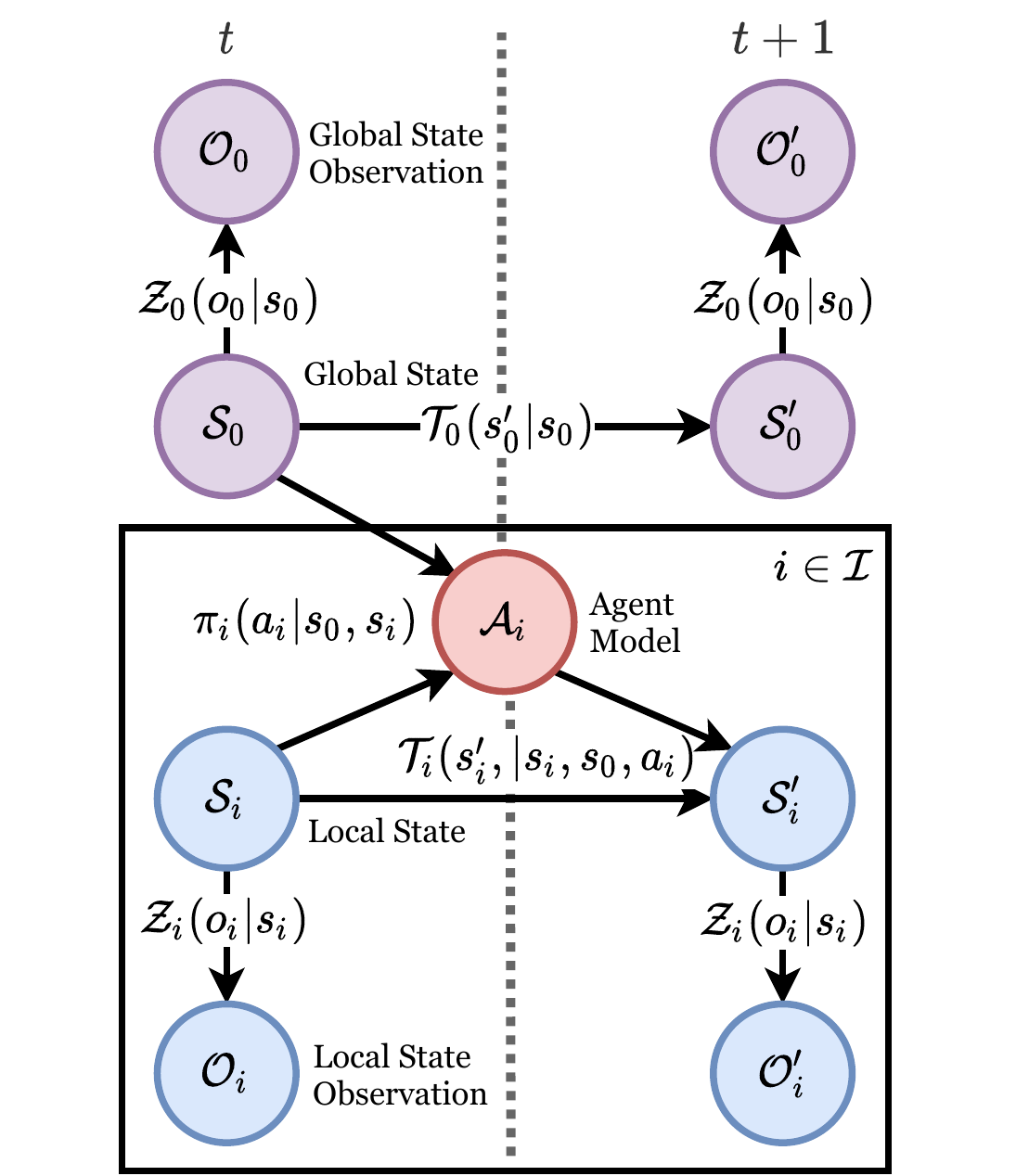}
    \caption{A DBN for transition-independent agent-aware state estimation.}
    \label{fig:dbn}
\end{figure}

To efficiently perform inference with the DBN, we can convert it into a factor graph and perform the sum-product message passing algorithm \cite{kschischang2001factor}. 

\begin{proposition}
    Inference over $t$ timesteps using $n$ agents in a TI-AASEP by applying the forward-backward sum-product message passing algorithm to the DBN has worst-case time complexity $O(tnk^2)$, where each node can take up to $k$ different discrete values.
    \label{prop:runtime-complexity}
\end{proposition}

\begin{proofsketch}
     The factor-graph representation of our $t$-timestep unrolled DBN has $O(tn)$ factors and $O(tn)$ variables, for each node and evidence-node respectively. Each variable performs at most $k$ additions, over at most a $k$-dimensional distribution, taking $O(k^2)$ time to summarize a constant number of incoming messages. As the factor graph is a tree of finite width, exactly one message needs to be passed per edge per direction for an exact solution. Therefore, the algorithm executes in $O(tnk^2)$ time, with a constant number of summarizations per variable. 
\end{proofsketch}

The worst-case time complexity $O(tnk^2)$ provides a desirable linear dependence on the number of agents in exchange for the more strict assumptions of a TI-AASEP. Intuitively, the worst-case time complexity of exact inference in the general AASEP case has a superlinear dependency on $n$. Thus, we see that the additional assumptions of transition-independent agents and local full observability for the agents improve the feasibility of runtime inference. 

In practice, we observe that clearly interacting or obstructed agents can be removed from inference. This justifies a locally fully observable transition-independent Dec-MDP, at the cost of not always using every observation. In the traffic scenario (Section~\ref{sec:intro}), this entails removing cars obstructed in their direction of travel. In the limit, removing every agent reduces our model to an HMM over the global state, a standard approach to traffic signal estimation. 

\begin{proposition}
    Inference of global state $s_0$ over $t$ timesteps using zero agents in a TI-AASEP is equivalent to inference of $s_0$ in an HMM, where $\mathcal{S}_0$ is the set of hidden states and $\mathcal{O}_0$ is the set of observations.
    \label{prop:degenhmm}
\end{proposition}

\begin{proofsketch}
   An HMM is a tuple $\langle \mathcal{S,T,O,Z} \rangle$, where $\mathcal{S}$ is a state space that transitions based on a transition function $\mathcal{T}$ and emits observations from a set $\mathcal{O}$ based on the observation function $\mathcal{Z}$. With no agents, a TI-AASEP only contains global state factors (all elements with a nonzero subscript disappear), and $\langle \mathcal{S}_0,\mathcal{T}_0, \mathcal{O}_0, \mathcal{Z}_0 \rangle$ forms an HMM. 
\end{proofsketch}

Proposition~\ref{prop:degenhmm} states that inference done with a TI-AASEP over no agents collapses to an HMM over the global state. As a result of the construction of the DBN, estimates of global state can only improve when agents are incorporated. Under minor violations of the TI-AASEP's assumptions, the TI-AASEP still often serves as a robust approximate model in practice. Finally, if fewer assumptions hold for a domain, it may be better to model the problem as a full AASEP.
 
\section{Traffic Light Classification}

To demonstrate the use of TI-AASEPs in practice, we apply it to the task of traffic light classification as discussed in Section~\ref{sec:intro} (Figure~\ref{fig:teaser}) for four-way intersections. This requires partially specifying a Dec-MDP, creating a TI-AASEP, and building the corresponding DBN, and results in accurate realtime inference in the classification problem.

We first construct a signalized, transition-independent locally fully observable Dec-MDP 
$\mathcal{M} = \langle \mathcal{I}, \mathcal{S}, \mathcal{A}, \mathcal{T}, \cdot, \cdot, \cdot \rangle$, 
where each observed, unobstructed vehicle at the intersection (excluding our own) is placed in the set of agents. The global state space $\mathcal{S}_0$ includes the possible traffic light states for the parallel and perpendicular directions of travel, i.e.\ $\{ \textsc{Red}, \textsc{Yellow}, \textsc{Green}\} \times \{ \textsc{Red}, \textsc{Yellow}, \textsc{Green}\}$. Illegal light combinations (e.g. Green, Green) are given 0 probability. Due to transition-independence, any agent that interacts with a pedestrian or other obstruction is removed from $\mathcal{I}$ before inference.

\begin{figure*}[t]
    \centering
    \includegraphics[height=2.5in]{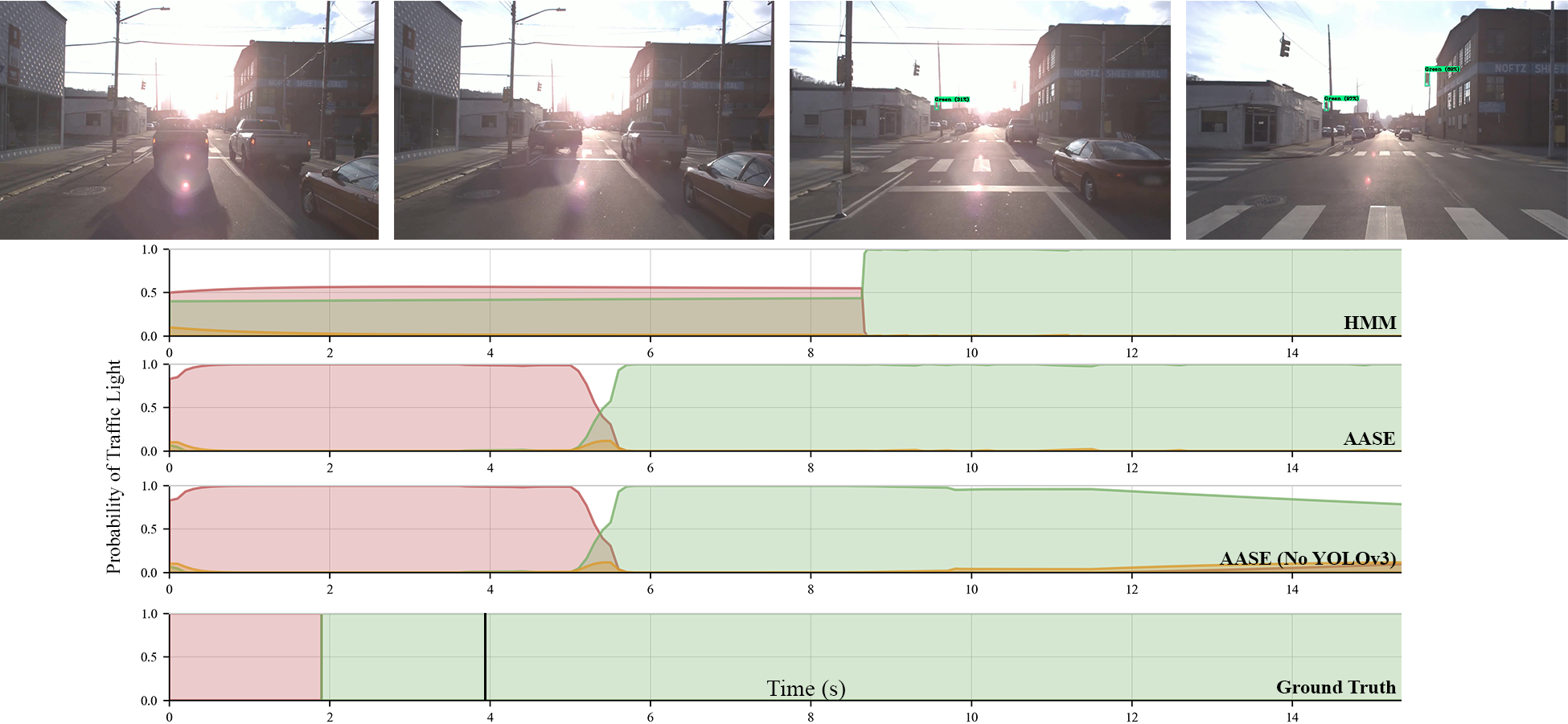}
    \caption{A typical scenario of an intersection where the light is obscured by glare. The ground truth transition from \emph{red} to \emph{green} occurs at 2 s, and a black line at 4 s denotes the first human reaction. The delay between the human and AASE is due to varying delays of the 3 cars at the intersection.}
    \label{fig:results}
\end{figure*}

Each agent $i \in \mathcal{I}$ has a corresponding local state space $\mathcal{S}_i$, which is the cross product of its position discretized as \{\textsc{AtIntersection}, \textsc{TurningLeft}, \textsc{DrivingStraight}, \textsc{TurningRight}\} and velocity discretized as \{\textsc{None} (0 m/s), \textsc{Low} (1-5 m/s), \textsc{High} ($>$5 m/s)\}. Additionally we split the agents into two groups, one for agents travelling in the same direction of travel and one for the perpendicular direction of travel. We represent actions as a cross product between its steering inputs \{\textsc{Left}, \textsc{Straight}, \textsc{Right}\} and accelerator inputs \{\textsc{Minus}, \textsc{Zero}, \textsc{Plus}\}. The transition probabilities $\mathcal{T}_i$ are specified via a simple approximate physics-based model for position and velocity, taking into account steering inputs and accelerator inputs in a straightforward manner. The transition probabilities $\mathcal{T}_0$ for the traffic signal are composed of a high probability self-transition, a low probability transition to the next light configuration in a predefined sequence, and a near zero probability for any out-of-sequence light changes (e.g.\ from $\textsc{Yellow}$ to $\textsc{Green}$).

The TI-AASEP requires a set of joint observations $\mathcal{O}$ representing perception of the light and other vehicles. Each $\mathcal{O}_{i>0}$ equals $\mathcal{S}_{i>0}$, as LiDAR was used to reconstruct estimates of position and velocity directly, which were then discretized. As vehicles can only directly observe the traffic light in their direction of travel, $\mathcal{O}_0 = \{ \textsc{Red}, \textsc{Yellow}, \textsc{Green}\}$. With these observations alone, the full global state cannot be reconstructed, as it includes a separate light for each direction of travel. We also defined a set of observation functions $\mathcal{Z}$ to describe sensor uncertainty and occlusions. We specified $\mathcal{Z}_0$ to be $92\%$ accurate and $\mathcal{Z}_{i>0}$ at $95\%$. The final component is $\Pi$, the set of policies that map local states to actions. Each agent policy $\pi_i$ is described by a driver model that assumes drivers are fairly stochastic but largely obey traffic laws. We can perform inference given the constructed DBN as structured in Figure~\ref{fig:dbn}.

\section{Experiments}

We compare AASE with a baseline HMM comparable to a current standard in traffic light detection and classification~\cite{gomez2014traffic}. Both approaches are implemented in Python using the library Pomegranate~\cite{schreiber2017pomegranate} for probabilistic models. Each model uses a convolutional neural network with a \emph{YOLOv3} architecture~\cite{yolov3}, trained on the \emph{Bosch Small Traffic Light data set}~\cite{behrendt2017deep}, as the 30 Hz vision detector that gives direct observations.  All trials are run on an Intel i7-6700k CPU at 4.0 GHz with an Nvidia RTX 2070 Super GPU. 

All experiments were run on a subset of the \emph{Argoverse 1.0 3D Tracking} dataset scenarios~\cite{chang2019tracking} that contained traffic lights. Argoverse is a dataset that includes both HD maps of multiple cities and sensor data collected by a real autonomous vehicle, along with ground truth vehicle and pedestrian positions as labels at 10 Hz. This rich combination of data is consumed by our multi-modal model, which uses raw pixel data, vehicle annotations, and HD maps to estimate the traffic light signals. From the \emph{Argoverse} data set, we chose 21 challenging scenarios that contain four-way intersections with traffic lights and hand-labeled the ground truth traffic light states. We provide these labels along with our source code at {\small \url{https://github.com/ikhatri/AASE}}.

Table~\ref{table:results} records accuracy under several artificial occlusion patterns. The columns of the table indicate occlusion lengths as percentages of the time that the traffic light was visible in each scenario. For all patterns, 0\% indicates no observations dropped and 100\% indicates all observations dropped. The HMM results with all traffic light observations dropped have been omitted, as the behavior collapses to predicting red no matter the scenario. Continuous(\texttt{start}) drops the first $x$\% of observations from the sequence, modeling a continuous occlusion at the start of the scenario. Continuous(\texttt{end}) drops the last $x$\% of observations, modeling an occlusion at the end of the scenario. Continuous(\texttt{random}) drops an $x\%$ length duration from the scenario from a random starting point. For each timestep, Discontinuous(\texttt{random}) drops each observation independently with $x\%$ probability. For both random patterns, we average over 5 trials, and report standard error. Accuracy is measured at 10 Hz via a 0--1 indicator loss between ground truth and the output sequence of most likely states (MLE) for each model, divided by the 3762 LiDAR frames in the data set. We observe that the runtime of AASE is roughly $(40 n + \alpha)$ ms, where $n$ is the number of agents and $\alpha$ is a constant overhead, empirically verifying Proposition~\ref{prop:runtime-complexity}.

\section{Discussion}

To illustrate our approach, we analyze a real-world intersection scenario in which the traffic light is completely occluded (Figure~\ref{fig:results}). In this scenario, the YOLOv3 model cannot detect the traffic light for the first 9 seconds due to glare from the sun. If the AV only used direct estimates of the light smoothed via an HMM to drive decision making, it would never have driven into the intersection, gained visibility of the traffic light, and discovered that the light was green. In contrast, by watching the other cars, AASE retains an accurate estimate of the light throughout the scenario, offset by a few seconds. In practice, other drivers' latency may add a human-like delay to AASE.

Table~\ref{table:results} shows the accuracy advantage that AASE has over an HMM during signal degradation events. We explicitly model two kinds of signal degradation: 1) continuous occlusions, which simulate prolonged obstructions of the traffic light and 2) discontinuous random patterns, which simulate noisy observations from the neural network. As degradation becomes severe, AASE uses observations of driver behavior to maintain an acceptable traffic light estimate.

There are clear differences between AASE and the HMM on each signal degradation pattern. On the continuous occlusion patterns, AASE smoothly decreases to $74\%$ as the duration of the temporally extended occlusions increase from $20\%$ to $100\%$ of each scenario. At $80\%$ occlusion, depending on the temporal nature of the pattern, the HMM is $14\%$ to $35\%$ worse than AASE in accuracy, well outside the margin of error. Moreover, in the case of continuous obstructions starting a scenario, the HMM even falls below $50\%$ accuracy and totally fails at $100\%$ occlusion.

\begin{table}[t]
    \scriptsize
    \caption{Classification accuracy (\%) for all intersection scenarios}
    \setlength{\tabcolsep}{2.7pt}
    \begin{tabularx}{\linewidth}{cccccccc}
        \toprule
        Occlusion & Method & 0\% & 20\% & 40\% & 60\% & 80\% & 100\%\\
        \midrule
        \multirow{2}{*}{Cont(\texttt{start})} & \textsc{AASE} & \textbf{90.2} & \textbf{86.4} & \textbf{83.0} & \textbf{80.0} & \textbf{75.4} & \textbf{74.0} \\
        & \textsc{HMM} & 88.1 & 76.9 & 64.5 & 51.7 & 40.5 & -- \\
        \midrule
        \multirow{2}{*}{Cont(\texttt{end})} & \textsc{AASE} & \textbf{90.2} & \textbf{88.7} & 86.5 & \textbf{83.8} & \textbf{81.9} & \textbf{74.0} \\
        & \textsc{HMM} & 88.1 & 86.4 & \textbf{88.5} & 73.0 & 67.6 & -- \\
        \midrule
        \multirow{2}{*}{Cont(\texttt{rand})} & \textsc{AASE} & \textbf{90.2} & \textbf{88.0$\boldsymbol{\pm}$0.6} & \textbf{85.8$\boldsymbol{\pm}$0.3} & \textbf{84.0$\boldsymbol{\pm}$0.1} & \textbf{77.7$\boldsymbol{\pm}$0.7} & \textbf{74.0} \\
        & \textsc{HMM} & 88.1 & 85.4$\pm$0.5 & 79.0$\pm$1.6 & 75.6$\pm$0.5 & 62.6$\pm$1.6 & -- \\
        \midrule
        \multirow{2}{*}{Discont(\texttt{rand})} & \textsc{AASE} & \textbf{90.2} & \textbf{89.7$\boldsymbol{\pm}$0.3} & \textbf{88.8$\boldsymbol{\pm}$0.2} & 87.5$\pm$0.2 & 84.9$\pm$0.4 & \textbf{74.0} \\
        & \textsc{HMM} & 88.1 & 87.7$\pm$0.4 & 87.4$\pm$0.6 & 87.1$\pm$0.6 & 85.9$\pm$0.9 & -- \\
        \bottomrule
    \end{tabularx}
    \label{table:results}
\end{table}

An asymmetry between performance of the HMM on starting versus ending obstructions is stark. This is due to sequential models relying heavily on the earliest observations to accurately initialize internal state, which, in the case of traffic lights, changes fairly infrequently. In fact, in Figure~\ref{fig:results}, we see that this real-world occluding glare example is of the more damaging obstruction starting pattern, and thus the HMM performs especially poorly, mirroring the simulations.

Finally, under the discontinuous noise pattern, both models remain accurate despite increasingly many independently dropped observations. This is because both models exploit an internal HMM structure that connects direct observations of the light to an estimate of the light state. Despite AASE using a more complex DBN model, the internal HMM is preserved and allows for it to be robust to random discontinuous noise. AASE does not trade off accuracy between continuous obstructions and discontinuous noise; rather, it improves or matches performance across a broad spectrum of scenarios.

We observe similar advantages to those above in the remaining scenarios, especially where the trained traffic light detection model systematically fails. The \emph{Bosch} data set used for training lacked both horizontal traffic lights and nighttime scenes, features present in a few of the \emph{Argoverse} data set scenarios. On these scenarios, the traffic light detection model performed poorly. This provides us with realistic failure profiles to compare AASE and HMM on their ability to recover. While a better data set may improve detection, a fleet of autonomous vehicles will likely be unable to totally avoid out of sample scenarios or occlusion~\cite{webb2020waymos}.

Our results show the benefits of using AASE to exploit a range of observations. It takes advantage of information often already collected (e.g.\ the position and behavior of other vehicles) and this additional information can often be collected with sensors that have failure modes independent of sensors used to directly observe the traffic lights. Methods such as LiDAR and radar are robust to glare, radar is robust in snow and rain, and even cameras are less likely to have multiple vehicles obstructed by glare or rain than the light itself. However, AASE is not simply sensor fusion; using AASE does not require pointing multiple sensors at the same object. For instance, we are using LiDAR observations of vehicles to update an estimate of the color of the traffic light, which would not be possible with conventional sensor fusion.

\section{Conclusion}

We propose the novel class of \emph{agent aware state estimation problems} (AASEPs) and present the \emph{agent-aware state estimation} framework to solve them. By modeling agents as though they are solving a signalized Dec-MDP, our framework can indirectly estimate state by observing the behavior of the agents in the Dec-MDP. Additionally we present a constrained form of AASEPs that enforces transition independence (TI-AASEPs), prove that an exact solution is linear with the number of agents, and that by construction the performance is superior to an HMM which ignores agents. Finally, we apply our TI-AASEP approach to traffic light classification to show empirically that our model retains a high accuracy in real-world occlusion scenarios where HMMs fail. Our approach combines various autonomous system capabilities, such as agent models and maps, to form an understanding of an intersection in real-time and improves state estimation in multi-agent systems.

\bibliographystyle{ieeetran}
\bibliography{bibliography}

\begin{thebibliography}{10}
\providecommand{\url}[1]{#1}
\csname url@rmstyle\endcsname
\providecommand{\newblock}{\relax}
\providecommand{\bibinfo}[2]{#2}
\providecommand\BIBentrySTDinterwordspacing{\spaceskip=0pt\relax}
\providecommand\BIBentryALTinterwordstretchfactor{4}
\providecommand\BIBentryALTinterwordspacing{\spaceskip=\fontdimen2\font plus
\BIBentryALTinterwordstretchfactor\fontdimen3\font minus
  \fontdimen4\font\relax}
\providecommand\BIBforeignlanguage[2]{{%
\expandafter\ifx\csname l@#1\endcsname\relax
\typeout{** WARNING: IEEEtran.bst: No hyphenation pattern has been}%
\typeout{** loaded for the language `#1'. Using the pattern for}%
\typeout{** the default language instead.}%
\else
\language=\csname l@#1\endcsname
\fi
#2}}

\bibitem{broggi2012vislab}
A.~Broggi, P.~Cerri, M.~Felisa, M.~C. Laghi, L.~Mazzei, and P.~P. Porta, ``{The
  {VisLab} Intercontinental Autonomous Challenge: An extensive test for a
  platoon of intelligent vehicles},'' \emph{IJVAS}, 2012.

\bibitem{svegliato2019belief}
J.~Svegliato, K.~H. Wray, S.~J. Witwicki, J.~Biswas, and S.~Zilberstein,
  ``{Belief Space Metareasoning for Exception Recovery},'' in \emph{IROS},
  2019.

\bibitem{basich2020learning}
C.~Basich, J.~Svegliato, K.~H. Wray, S.~Witwicki, J.~Biswas, and
  S.~Zilberstein, ``{Learning to Optimize Autonomy in Competence-Aware
  Systems},'' \emph{arXiv preprint arXiv:2003.07745}, 2020.

\bibitem{achtelik2012sfly}
M.~Achtelik, M.~Achtelik, Y.~Brunet, M.~Chli, S.~Chatzichristofis,
  \emph{et~al.}, ``{Sfly: Swarm of micro flying robots},'' in \emph{IROS},
  2012.

\bibitem{cliff2015online}
O.~M. Cliff, R.~Fitch, S.~Sukkarieh, D.~L. Saunders, and R.~Heinsohn, ``{Online
  localization of radio-tagged wildlife with an autonomous aerial robot
  system},'' in \emph{RSS}, 2015.

\bibitem{goodrich2008supporting}
M.~A. Goodrich, B.~S. Morse, D.~Gerhardt, J.~L. Cooper, M.~Quigley, J.~A.
  Adams, and C.~Humphrey, ``{Supporting wilderness search and rescue using a
  camera-equipped mini {UAV}},'' \emph{JFR}, 2008.

\bibitem{pineda2016continual}
L.~Pineda, T.~Takahashi, H.-T. Jung, S.~Zilberstein, and R.~Grupen,
  ``{Continual planning for search and rescue robots},'' in \emph{Humanoids},
  Seoul, Korea, 2015.

\bibitem{wray2016hierarchical}
K.~H. Wray, L.~Pineda, and S.~Zilberstein, ``{Hierarchical Approach to Transfer
  of Control in Semi-Autonomous Systems},'' in \emph{IJCAI}, 2016.

\bibitem{zilberstein2015building}
S.~Zilberstein, ``{Building strong semi-autonomous systems},'' in \emph{AAAI},
  2015.

\bibitem{kaelbling1998planning}
L.~P. Kaelbling, M.~L. Littman, and A.~R. Cassandra, ``{Planning and acting in
  partially observable stochastic domains},'' \emph{AIJ}, 1998.

\bibitem{goldman2003optimizing}
C.~V. Goldman and S.~Zilberstein, ``{Optimizing information exchange in
  cooperative multi-agent systems},'' in \emph{AAMAS}, 2003.

\bibitem{vaseghi1997noise}
S.~V. Vaseghi and B.~P. Milner, ``{Noise compensation for hidden {M}arkov model
  speech recognition in adverse environments},'' \emph{IEEE Transactions on
  SAP}, 1997.

\bibitem{ghahramani2001introduction}
Z.~Ghahramani, ``{An introduction to hidden {M}arkov models and Bayesian
  networks},'' in \emph{Hidden Markov models: Applications in computer
  vision}.\hskip 1em plus 0.5em minus 0.4em\relax World Scientific, 2001.

\bibitem{rabiner1986introduction}
L.~Rabiner and B.~Juang, ``{An introduction to hidden {M}arkov models},''
  \emph{IEEE AASP}, 1986.

\bibitem{dymarski2011hidden}
P.~Dymarski, \emph{{Hidden {M}arkov Models: Theory and Applications}}.\hskip
  1em plus 0.5em minus 0.4em\relax BoD Books on Demand, 2011.

\bibitem{gomez2014traffic}
A.~Gomez, F.~Alencar, P.~Prado, F.~Osorio, and D.~Wolf, ``{Traffic lights
  detection and state estimation using {HMMs}},'' in \emph{IEEE IV}, 2014.

\bibitem{murphy2002dynamic}
K.~P. Murphy, ``{Dynamic bayesian networks: representation, inference and
  learning},'' \emph{PhD Thesis}, 2002.

\bibitem{oliehoek2016concise}
F.~A. Oliehoek, C.~Amato, \emph{et~al.}, \emph{{A concise introduction to
  decentralized POMDPs}}.\hskip 1em plus 0.5em minus 0.4em\relax Springer,
  2016.

\bibitem{mundhenk2000complexity}
M.~Mundhenk, J.~Goldsmith, C.~Lusena, and E.~Allender, ``Complexity of
  finite-horizon markov decision process problems,'' \emph{Journal of the ACM
  (JACM)}, vol.~47, no.~4, pp. 681--720, 2000.

\bibitem{bernstein2002complexity}
D.~S. Bernstein, R.~Givan, N.~Immerman, and S.~Zilberstein, ``{The complexity
  of decentralized control of {M}arkov decision processes},'' \emph{MOR}, 2002.

\bibitem{amato2013decentralized}
C.~Amato, G.~Chowdhary, A.~Geramifard, N.~K. {\"U}re, and M.~J. Kochenderfer,
  ``{Decentralized control of partially observable {M}arkov decision
  processes},'' in \emph{CDC}.\hskip 1em plus 0.5em minus 0.4em\relax IEEE,
  2013.

\bibitem{SZ:BZLGjair04}
R.~Becker, S.~Zilberstein, V.~Lesser, and C.~V. Goldman, ``{Solving Transition
  Independent Decentralized {MDPs}},'' \emph{JAIR}, 2004.

\bibitem{kschischang2001factor}
F.~R. Kschischang, B.~J. Frey, and H.-A. Loeliger, ``{Factor graphs and the
  sum-product algorithm},'' \emph{IEEE Transactions on IT}, 2001.

\bibitem{schreiber2017pomegranate}
J.~Schreiber, ``{Pomegranate: Fast and flexible probabilistic modeling in
  {P}ython},'' \emph{JMLR}, 2017.

\bibitem{yolov3}
J.~Redmon and A.~Farhadi, ``{{YOLO}v3: An Incremental Improvement},''
  \emph{arXiv:1804.02767}, 2018.

\bibitem{behrendt2017deep}
K.~Behrendt, L.~Novak, and R.~Botros, ``{A deep learning approach to traffic
  lights},'' in \emph{ICRA}, 2017.

\bibitem{chang2019tracking}
M.-F. Chang, J.~Lambert, P.~Sangkloy, J.~Singh, S.~Bak, A.~Hartnett, D.~Wang,
  P.~Carr, S.~Lucey, D.~Ramanan, and J.~Hays, ``{Argoverse: {3D} Tracking and
  Forecasting With Rich Maps},'' in \emph{CVPR}, 2019.

\bibitem{webb2020waymos}
N.~Webb, D.~Smith, C.~Ludwick, T.~Victor, Q.~Hommes, F.~Favaro, G.~Ivanov, and
  T.~Daniel, ``Waymo's safety methodologies and safety readiness
  determinations,'' 2020.

\end{thebibliography}

\end{document}